%% file: main.tex
\documentclass[journal]{IEEEtran}
\usepackage{cite}
\usepackage{amsmath,amssymb,amsfonts}
\usepackage{graphicx}
\usepackage{textcomp}
\usepackage{xcolor}
\usepackage[utf8]{inputenc}
\usepackage{multirow}
\usepackage{tikz}
\usepackage{pgfplots}
\pgfplotsset{compat=1.17} 
\usepgfplotslibrary{groupplots}
\usepackage{tabularx}
\usepackage{hyperref}

\hypersetup{
    colorlinks=false,
    pdfborder={0 0 0},
    hidelinks
}

\usepackage{algorithm}
\usepackage{algorithmic}

\usepackage{booktabs}
\usepackage{siunitx}
\usepackage{subcaption} 
\usepackage{mathtools}

\DeclareMathOperator*{\E}{\mathbb{E}}
\DeclareMathOperator{\Var}{Var}
\DeclareMathOperator{\Concat}{Concat}

\def\BibTeX{{\rm B\kern-.05em{\sc i\kern-.025em b}\kern-.08em
    T\kern-.1667em\lower.7ex\hbox{E}\kern-.125emX}}
\begin{document}

\title{DeepFedNAS: Efficient Hardware-Aware Architecture Adaptation for Heterogeneous IoT Federations via Pareto-Guided Supernet Training
\thanks{The computations in this work were enabled by resources provided by the National Academic Infrastructure for Supercomputing in Sweden (NAISS), partially funded by the Swedish Research Council through grant agreement no. 2022-06725.}
\thanks{A preliminary version of this work has been accepted at the ESANN 2026 conference \cite{Khan7335}. This manuscript significantly extends the conference version with new theoretical analysis, a hardware-aware optimization framework, additional experiments, and deeper discussion.}
}

\author{Bostan~Khan and Masoud~Daneshtalab
\IEEEcompsocitemizethanks{
\IEEEcompsocthanksitem B. Khan and M. Daneshtalab are with M\"{a}lardalen University, V\"{a}ster\r{a}s, Sweden.\protect\\
E-mail: \{bostan.khan, masoud.daneshtalab\}@mdu.se}
}

\maketitle

\begin{abstract}
Deploying federated learning across heterogeneous IoT device fleets requires tailored neural network architectures for each device class, yet existing Federated Neural Architecture Search (FedNAS) methods suffer from unguided supernet training and prohibitively costly post-training search pipelines that validate thousands of subnets to construct learned accuracy predictors. We introduce \textbf{DeepFedNAS}, a two-phase framework built on a multi-objective fitness function that synthesizes information-theoretic network metrics with architectural heuristics. In the first phase, \emph{Federated Pareto Optimal Supernet Training} replaces random subnet sampling with a pre-computed cache of elite, high-fitness architectures, yielding a superior supernet. In the second phase, a \emph{Predictor-Free Search} uses the structural fitness function as an accuracy proxy without constructing a learned subnet-accuracy predictor. In our CIFAR-10 benchmark, preparing the baseline predictor requires evaluating 10,000 subnets over the 5,000-image validation split, totaling 50 million image-level forward evaluations. DeepFedNAS eliminates these evaluations and selects a hardware-optimized architecture in $\sim$20 seconds on a CPU. Experiments on CIFAR-10, CIFAR-100, and CINIC-10 demonstrate state-of-the-art accuracy and robust performance under extreme non-IID conditions ($\alpha=0.1$). On CIFAR-100, DeepFedNAS provides an average 2.12-percentage-point gain across the four computation-budget intervals. Under the lowest evaluated computation budget, its mean result exceeds SuperFedNAS's best mean accuracy while using $2.95\times$ fewer parameters. These results make DeepFedNAS practical for scalable, communication-constrained IoT federations.
Source code: \url{https://github.com/bostankhan6/DeepFedNAS}
\end{abstract}


\input{sections/01_introduction}
\input{sections/02_related_work}
\input{sections/03_methodology}
\input{sections/04_experiments}
\input{sections/05_limitations_and_future_work}
\input{sections/06_conclusion}
\input{sections/backmatter}

%% file: sections/01_introduction.tex
\section{Introduction}
\label{sec:introduction}

The rapid expansion of the Internet of Things (IoT) has created networks of billions of devices with profoundly heterogeneous hardware profiles, spanning resource-rich edge gateways and smart cameras, mid-range wearable health monitors, and severely constrained sensor nodes \cite{nguyen2021fl_iot_survey}. Deploying intelligent inference models across such diverse device fleets requires architectures that are individually tailored to each device's computational, memory, and energy budget. Federated Learning (FL) \cite{mcmahan2017communication} has emerged as the dominant paradigm for training these models collaboratively across decentralized devices without exposing private data, enabling applications ranging from mobile keyboard prediction \cite{hard2018federated} to personalized healthcare analytics \cite{silva2019federated}. While FL provides a robust solution for \emph{how} to train a given model, the challenge of determining \emph{what} model architecture to deploy on each device class remains a critical operational bottleneck. This architectural design, often performed manually, is frequently suboptimal given the vast heterogeneity of data distributions and hardware capabilities across IoT devices \cite{li2020federated}.

In communication-constrained IoT federations, this challenge is compounded by two practical realities. First, every model parameter transmitted during FL training directly impacts bandwidth consumption and energy expenditure which is a significant concern for devices communicating over metered cellular connections (LTE-M, NB-IoT) or low-power wireless links \cite{mills2022comms_efficient_fl_iot}. Deploying unnecessarily large models wastes this scarce bandwidth. Second, IoT device fleets are not static: new hardware platforms, firmware revisions, and deployment contexts emerge continuously, requiring deployment pipelines that can rapidly produce optimized architectures for new device classes without costly retraining \cite{imteaj2022fl_constrained_iot}. These operational demands motivate the need for automated architecture search methods that are both communication-efficient and rapidly adaptable.

To automate this critical design step, the research community has developed Federated Neural Architecture Search (FedNAS) to discover optimal network architectures directly within the federated setting \cite{he2020towards}. Among the various FedNAS techniques, supernet-based methods have become the state-of-the-art due to their high cost efficiency, building on the success of centralized approaches like the Once-For-All (OFA) network \cite{cai2020once}. The SuperFedNAS framework \cite{khare2023superfednas}, in particular, marked a key advancement by decoupling the costly supernet training phase from a fast, training-free search phase. This innovation effectively reduced the training complexity of finding specialized subnets for multiple hardware targets from $O(N)$ to $O(1)$.

Despite this progress, the SuperFedNAS framework presents two fundamental challenges that limit its ultimate performance and practicality for IoT deployments. First, the \emph{training of the supernet itself is unguided and inefficient}. The baseline framework predominantly relies on random sampling of subnets (e.g., the Sandwich Rule's uniform sampling component \cite{Yu2019ICCV}) to update the shared weights. This unguided approach frequently selects suboptimal architectures, whose subsequent training provides noisy and conflicting gradient updates that can degrade the supernet's shared weights and hinder the performance of higher-performing subnets, ultimately wasting scarce communication bandwidth between IoT devices and the aggregation server. Second, the \emph{post-training search for optimal subnets remains a computationally expensive, multi-stage pipeline}. The standard methodology generates a large dataset of architecture--accuracy pairs through subnet evaluation and fits a separate accuracy predictor before search. In our CIFAR-10 benchmark, constructing this predictor requires 10,000 complete validation-set subnet evaluations, corresponding to 50 million image-level forward evaluations. This creates a substantial GPU-dependent post-training bottleneck.

In this paper, we introduce \textbf{DeepFedNAS} (illustrated in Fig.~\ref{fig:deepfednas_pipeline}), a novel, two-phase framework that comprehensively addresses these limitations to significantly advance federated supernet training and search. Our work is grounded in the observation that a mathematically motivated approach to architecture evaluation can fundamentally transform both the supernet training process and the subsequent search for optimal subnets. Inspired by the mathematical design concepts of DeepMAD \cite{Shen2023CVPR}, we construct a unified multi-objective fitness function, $\mathcal{F}(\mathcal{A})$, that synthesizes network information theory with empirical architectural heuristics. Crucially, we operationalize structural guidelines (like depth uniformity and channel monotonicity) as penalty terms, thereby enabling a holistic, single-objective optimization for all desired architectural properties within a federated, weight-sharing context. This unified fitness function underpins our two primary contributions.

First, to address the supernet training inefficiency, we propose \emph{Federated Pareto Optimal Supernet Training}. Leveraging our fitness function, we perform an extensive, independent search to generate a pre-computed cache of Pareto-optimal architectures, which we term the ``Pareto path''. This path serves as a Pareto-Guided training curriculum. Instead of random sampling, our methodology ensures the supernet's shared weights are consistently updated by gradients from theoretically grounded, high-fitness architectures, thereby producing a superior final supernet. Additionally, to fully enable this guided search and overcome inherent limitations of prior supernet implementations, we introduce a \emph{redesigned, generic ResNet-based~\cite{resnet} supernet framework}  that significantly expands the architectural search space and improves model fitness.

Second, we leverage the properties of our optimal path-guided trained supernet to introduce a \emph{Predictor-Free Search Method}\footnote{In this work, \emph{Predictor-Free} specifically denotes the absence of a learned subnet-accuracy predictor. For latency-constrained deployment, we use a lightweight latency predictor whose per-architecture label collection uses single-input profiling over a small number of repeated runs and is therefore substantially less costly than full validation-set accuracy evaluation, as detailed in Section~\ref{subsubsec:inference_latency}.}. After our supernet is trained using the path-guided curriculum, its shared weights are exceptionally well-conditioned for promising architectural patterns. We demonstrate that our mathematical fitness function, $\mathcal{F}(\mathcal{A})$, now serves as a low-cost architecture-based proxy for subnet accuracy that requires neither subnet training nor validation-set evaluation. Consequently, we can find optimal architectures for new MAC or parameter budgets by running a fast, on-demand genetic algorithm that maximizes this fitness function directly; latency-constrained deployment additionally uses the lightweight device-specific latency predictor described in Section~\ref{subsubsec:inference_latency}. This completely obviates the need for the costly, data-driven accuracy-predictor pipeline used in prior work. Critically, the fitness function itself is computable without executing the model on the target device.

Our comprehensive DeepFedNAS framework not only achieves state-of-the-art accuracy across diverse datasets and non-IID conditions but also dramatically accelerates the post-training search. We replace a multi-hour predictor pipeline with a rapid, on-demand search for specialized subnets adaptable to explicit hardware constraints like parameters and real-world latency.

In summary, our main contributions are:
\begin{itemize}
    \item A unified, multi-objective fitness function for CNN-based subnets that adapts information-theoretic network design concepts \cite{Shen2023CVPR} into a single searchable metric balancing expressiveness, stability, and structural soundness within a federated, weight-sharing context.
    \item A \emph{Federated Pareto Optimal Supernet Training} methodology, enabled by a redesigned generic ResNet-based supernet with a significantly expanded search space, which replaces unguided random sampling with a pre-computed cache of elite architectures as an intelligent training curriculum.
    \item A \emph{Predictor-Free Search Method} that uses this fitness function as a low-cost architecture-based proxy for subnet accuracy, eliminating the costly accuracy-predictor pipeline and enabling on-demand, hardware-aware subnet discovery in $\sim$20 seconds, allowing rapid adaptation to new IoT device classes without supernet retraining or accuracy-predictor construction.
    \item Extensive validation demonstrating state-of-the-art accuracy across three datasets and three data-heterogeneity settings, together with hardware-aware deployment optimization for heterogeneous edge devices. On CIFAR-100, under the lowest evaluated computation budget, the mean DeepFedNAS result exceeds SuperFedNAS's best mean accuracy while using $2.95\times$ fewer parameters.
\end{itemize}

%% file: sections/02_related_work.tex
\section{Related Work}
\label{sec:related_work}

\subsection{Federated Learning}
FL enables distributed learning without centralizing private data \cite{LIU2024128019}. FedAvg \cite{mcmahan2017communication} established the standard client-training/server-aggregation paradigm \cite{kairouz2019advances, li2019federatedchallenges, cooray2025systematicreview}. Our work centers on \emph{statistical heterogeneity}, where non-IID data degrades performance \cite{konevcny2016federated, FedNonIIDData}, and \emph{system heterogeneity}, where client resources differ \cite{li2020federated}, motivating models adapted to both local distributions \cite{makhija2024bayesian, tan2021fedproto} and hardware budgets \cite{farcas2022elasticity}.

\subsection{Federated Learning for IoT and Edge Intelligence}
\label{subsec:fl_iot}
FL is attractive for distributed IoT intelligence, but IoT deployments intensify resource and communication heterogeneity \cite{nguyen2021fl_iot_survey, khan2021fl_iot_challenges}. Devices range from GPU-equipped cameras to microcontroller-class sensors \cite{imteaj2022fl_constrained_iot}, while bandwidth-limited links make transmitted model size a practical bottleneck \cite{mills2022comms_efficient_fl_iot}. These conditions favor methods that train once yet expose architectures for different device classes. Prior work has used knowledge distillation \cite{lin2020ensemble_distill_fl} and pruning \cite{horvath2021fjord}; DeepFedNAS instead targets efficient supernet training and rapid hardware-aware specialization.

\subsection{Neural Architecture Search with Weight Sharing}
NAS automates architecture design. Early reinforcement-learning \cite{zoph2016nasrl} and evolutionary \cite{liu2017progressive} methods were costly, while weight-sharing supernets \cite{bender2018understanding, Weight-Sharing} and DARTS \cite{liu2018darts} improved efficiency. OFA \cite{cai2020once} supports multiple hardware constraints from one supernet, while BigNAS \cite{bignas} popularized the ``Sandwich Rule'' \cite{Yu2019ICCV}. Shared weights can nevertheless yield poor rank correlation with standalone accuracy \cite{pan2022distribution}, partly due to multi-model forgetting \cite{zhang2020forgetting, ma2025orthogonal}, motivating improved subnet sampling \cite{venkatesha2023dcnas,wang2023prenas}.

\subsection{Federated Neural Architecture Search}
FedNAS combines FL and NAS under decentralized data and hardware constraints \cite{he2020towards, garg2020direct, yu2022resourceaware}. Existing work targets either data personalization or system efficiency. Data-personalized methods adapt subnet paths or structures to local distributions \cite{roth2021federated, yan2024peaches, hoang2021personalized, yao2021fedmodelsearchrl, yao2024perfedrlnas, liu2025collaborativenas, yang2025hapfnas}. System-oriented methods expose differently sized subnets for heterogeneous clients \cite{horvath2021fjord}, but tightly coupled search and training can remain costly \cite{wang2025energy, liu2024finch}, and restricted search spaces can limit deployment diversity \cite{yu2022resourceaware}.

SuperFedNAS \cite{khare2023superfednas} addresses search cost by decoupling supernet training from post-training architecture search. Nevertheless, it still relies on unguided random subnet sampling and a learned accuracy predictor for post-training search. DeepFedNAS targets both stages: fitness-guided sampling improves supernet training, while the same structural fitness function supports Predictor-Free Search. This direction is related to centralized DC-NAS \cite{venkatesha2023dcnas}, which diversifies sampling, and PreNAS \cite{wang2023prenas}, which uses a zero-cost proxy to emphasize high-quality candidates.

\subsection{Mathematical Frameworks for Architecture Design}
Most NAS methods optimize architectures through learned or empirical search signals, including gradients \cite{liu2018darts}, evolution \cite{Real_Aggarwal_Huang_Le_2019}, and more recently LLM-derived priors \cite{fang2025llmgnas, qin2024flnas_llm}. DeepMAD \cite{Shen2023CVPR} instead uses intrinsic mathematical metrics related to network information entropy and effectiveness, while other approaches use mutual-information analysis \cite{namekawa2021mianas}. DeepFedNAS adapts this theory-driven perspective to federated supernet training by combining mathematical metrics with structural heuristics in a unified fitness function used for both guided training and Predictor-Free Search.

%% file: sections/03_methodology.tex
\section{Methodology}
\label{sec:methodology}

\begin{figure*}[t]
    \centering
    \includegraphics[width=\textwidth]{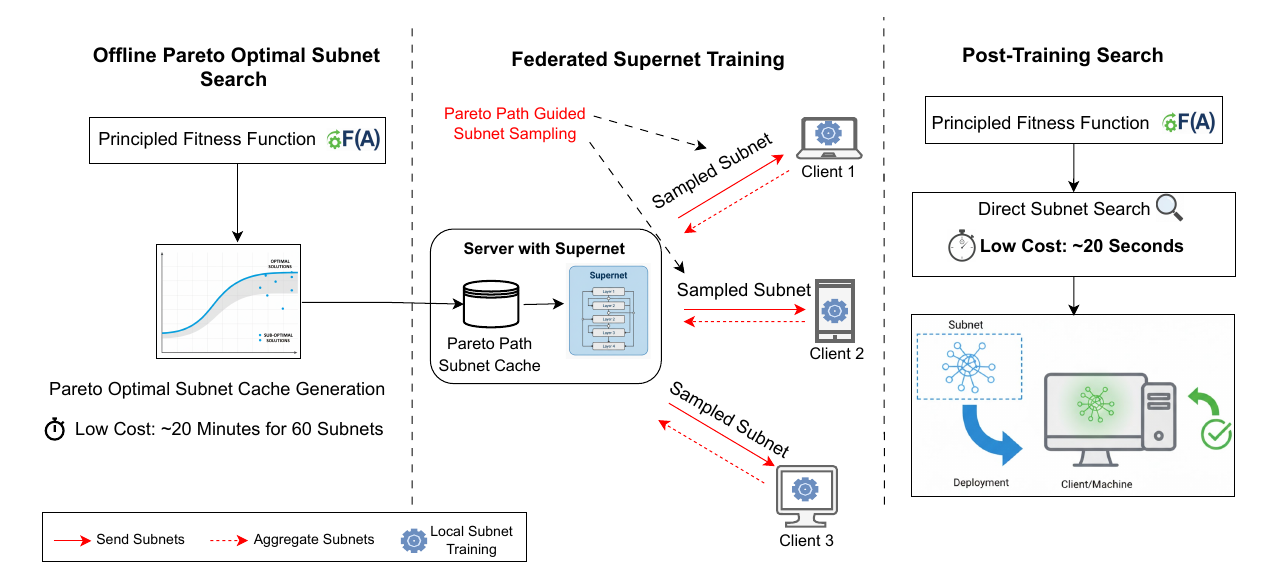} 
    \caption{\textbf{DeepFedNAS Pipeline.} This diagram illustrates the three operational stages that implement the two main phases of our framework. First, the \textbf{Offline Pareto Optimal Subnet Search} generates a ``Pareto Path Subnet Cache" of high-fitness architectures (e.g., 60 subnets in $\sim$20 minutes) using a multi-objective fitness function $\mathcal{F}(\mathcal{A})$. Second, the \textbf{Federated Supernet Training} leverages this cache for ``Pareto Path Guided Subnet Sampling," where clients are assigned highly effective architectures for local training, improving supernet weight quality. Finally, the \textbf{Post-Training Search} directly optimizes against $\mathcal{F}(\mathcal{A})$ to find an optimal subnet for deployment (e.g., in $\sim$20 seconds), completely bypassing the costly post-training accuracy-predictor pipeline. This is a critical advantage for heterogeneous IoT deployments where new device classes emerge frequently.}
    \label{fig:deepfednas_pipeline}
\end{figure*}

DeepFedNAS is organized into two main phases, implemented through three connected operational stages. Phase 1, Federated Pareto Optimal Supernet Training, consists of an offline architecture search that constructs a cache of high-fitness architectures (Section~\ref{subsec:offline_search}), followed by federated supernet training using this cache as a subnet-assignment curriculum (Section~\ref{subsec:path_guided_training}). Phase 2, Predictor-Free Search, ranks candidate deployment architectures using the same structural fitness function (Section~\ref{subsec:predictor_free}). The architecture-only structural fitness function underlying both phases is defined in Section~\ref{subsec:fitness_function}. We also use a redesigned generic ResNet-based supernet (Section~\ref{subsec:supernet_design}) to support the search space considered in this work. Figure~\ref{fig:deepfednas_pipeline} summarizes these stages. We begin by defining the federated supernet training problem.

\subsection{Problem Formulation: Federated Supernet Training}
\label{subsec:problem_formulation}

\paragraph{System model.}
We consider cross-device federated learning with a server and $K$ decentralized clients. Client $k \in \{1,\ldots,K\}$ holds a private local dataset $\mathcal{D}_k$, and $\mathcal{D}=\bigcup_{k=1}^{K}\mathcal{D}_k$ denotes the aggregate dataset. The server maintains a weight-sharing supernet with parameters $W$ and architectural search space $\mathbb{S}$. An architecture $\mathcal{A}\in\mathbb{S}$ identifies a subnet, and $\mathcal{G}(W,\mathcal{A})$ extracts its active parameters from the supernet. In round $t$, the server selects a client set $\mathcal{K}_t$ and assigns an architecture $\mathcal{A}_k$ to each $k\in\mathcal{K}_t$. After training, a separate deployment search selects an architecture $\mathcal{A}_{\mathrm{deploy}}$ for a requested MAC, parameter, or latency budget.

The weight-training objective is to learn shared weights $W$ that serve the architectural search space across the client data distributions. Let $\boldsymbol{\mathcal{A}}=(\mathcal{A}_1,\ldots,\mathcal{A}_K)$ denote the client-specific architecture assignments drawn from the training assignment distribution $P_{\mathrm{assign}}$:
\begin{equation}
\label{eq:fl_objective}
\underset{W}{\min} \; \E_{\boldsymbol{\mathcal{A}} \sim P_{\mathrm{assign}}} \left[ \sum_{k=1}^{K} \frac{|\mathcal{D}_k|}{|\mathcal{D}|} \mathcal{L}_k(\mathcal{G}(W, \mathcal{A}_k)) \right]
\end{equation}
where $\mathcal{L}_k$ is the empirical loss on client $k$'s local data, and the expectation is over the client-specific architecture assignments used during supernet training. Equation~\ref{eq:fl_objective} optimizes the shared supernet weights $W$ through local data-dependent updates. It is distinct from the structural architecture-selection objective introduced in Section~\ref{subsec:fitness_function}, which ranks candidate architectures $\mathcal{A}$.

The training process involves the following multi-step procedure in each round $t$:
\begin{enumerate}
    \item A subset of available clients, $\mathcal{K}_t$, is selected.
    \item For each client $k \in \mathcal{K}_t$, a subnet architecture $\mathcal{A}_k \in \mathbb{S}$ is assigned. The corresponding weights $w_{k,t} = \mathcal{G}(W_t, \mathcal{A}_k)$ are extracted and sent.
    \item Each client trains $w_{k,t}$ on $\mathcal{D}_k$, producing updated weights $w'_{k,t}$.
    \item The server receives updates and performs an overlap-aware aggregation to update $W_{t+1}$.
\end{enumerate}

The \textbf{SuperFedNAS framework \cite{khare2023superfednas}} serves as our baseline, specifically for its aggregation mechanism. To handle the sparse updates inherent in supernet training (Step 4), we employ \textbf{MaxNet} with an overlap-aware scaling factor. We refine the formalization of this aggregation using precise binary mask notation to account for parameter overlap. Let $\mathbb{I}_k(\theta) \in \{0,1\}$ indicate if a specific parameter $\theta$ is active in the architecture $\mathcal{A}_k$ assigned to client $k$. The server update $\Delta_\theta$ at round $t$ is computed as:
\begin{equation}
\label{eq:maxnet_agg}
\Delta_\theta = \frac{ \beta_t \mathbb{I}_{k_{\max}}(\theta) \Delta_{k_{\max}}(\theta) + (1-\beta_t) \sum_{k \in \mathcal{K}'} \mathbb{I}_k(\theta) \Delta_k(\theta) }{ \beta_t \mathbb{I}_{k_{\max}}(\theta) + (1-\beta_t) \sum_{k \in \mathcal{K}'} \mathbb{I}_k(\theta) + \epsilon }
\end{equation}
where $\mathcal{K}' = \mathcal{K}_t \setminus \{k_{\max}\}$ denotes the set of clients excluding the one training the largest subnet. The term $\epsilon$ is a small constant added for numerical stability. Here, $\Delta_k(\theta) = w'_{k,t}(\theta) - w_{k,t}(\theta)$ is the local update, $k_{\max}$ is the client assigned the largest subnet ($\mathcal{A}_{\max}$), and $\beta_t \in [0, 1]$ is a dynamic weighting coefficient that anneals over time using cosine scheduling. This aggregation ensures that shared weights are updated proportionally to their usage frequency. The global model is then updated as $W_{t+1}(\theta) = W_t(\theta) + \eta_g \cdot \Delta_\theta$, where $\eta_g$ denotes the server-side update step size.

The critical challenge lies in the sampling strategy (Step 2). SuperFedNAS's random sampling component is unguided. The architectural search space $\mathbb{S}$ is not only vast but also qualitatively uneven; the overwhelming majority of possible architectures are mediocre. This leads to inefficient gradient updates and a limited training curriculum. This fundamental limitation motivates our work.

\subsection{Supernet Framework and Search Space Definition}
\label{subsec:supernet_design}

Our initial investigation revealed that the baseline SuperFedNAS search space was inherently constrained, yielding architectures with consistently low fitness scores (Section~\ref{subsec:fitness_function}), as clearly illustrated by the limitations of the baseline's constrained search space and unguided random sampling shown in Fig.~\ref{fig:search_comparison}. This motivated a redesigned, generic ResNet-based supernet framework. This new framework is highly configurable, supporting a variable number of stages ($S$) and fine-grained control over architectural choices within each stage. We measure computational cost using multiply-accumulate operations (MACs), where one GigaMAC (GMAC) equals $10^9$ MACs.

\begin{figure}[h]
\centering
\resizebox{\columnwidth}{!}{%
\begin{tikzpicture}
\begin{axis}[
    title={Comparison of Network Fitness vs. Computational Cost},
    xlabel={GigaMACs (GMACs)},
    ylabel={Network Fitness Score},
    xmin=0, xmax=3.5,
    ymin=800, ymax=2500,
    legend pos=south east,
    ymajorgrids=true,
    xmajorgrids=true,
    grid style=dashed,
]

\addplot+[
    color=blue,
    mark=*,
    only marks,
    mark size=1.5pt
]
coordinates {
    (0.4582,1996.4345) (0.5084,2011.0045) (0.5587,2033.3150) (0.6085,2061.2231) (0.6585,2066.9836) (0.7051,2107.3541) (0.7581,2129.8347) (0.8083,2135.4749) (0.8552,2143.3050) (0.9078,2150.4828) (0.9555,2172.7109) (1.0074,2174.3215) (1.0559,2194.8507) (1.1071,2212.7052) (1.1561,2221.3589) (1.2038,2245.5879) (1.2554,2243.5945) (1.2966,2247.8605) (1.3557,2268.5116) (1.4071,2262.7110) (1.4556,2273.2589) (1.5055,2285.8887) (1.5524,2293.4120) (1.6017,2299.1613) (1.6567,2311.8298) (1.7061,2315.2586) (1.7563,2324.1325) (1.7984,2323.1748) (1.8504,2332.4577) (1.9026,2331.4937) (1.9536,2340.7370) (2.0006,2343.4821) (2.0518,2347.8205) (2.1058,2356.7579) (2.1531,2354.3407) (2.2048,2363.6753) (2.2520,2365.9564) (2.3039,2369.8908) (2.3511,2372.6093) (2.3983,2374.9003) (2.4549,2378.5600) (2.4998,2381.5707) (2.5540,2384.5180) (2.6036,2387.3436) (2.6508,2389.8922) (2.6956,2392.3798) (2.7522,2395.2490) (2.7994,2397.4598) (2.8466,2400.1490) (2.9032,2402.6661) (2.9504,2405.0198) (2.9976,2407.2014) (3.0448,2408.9382) (3.1014,2411.7059) (3.1486,2413.7511) (3.1958,2415.4171) (3.2524,2418.2864) (3.2996,2419.8599) (3.3279,2421.2482) (3.3656,2422.6366)
};
\addlegendentry{Pareto Path (Ours)}

\addplot+[
    color=red,
    mark=*,
    only marks,
    mark size=1.pt
]
coordinates {
    (0.5651,922.3063) (0.5769,922.5801) (0.5793,921.8258) (0.6076,927.3097) (0.6312,929.0074) (0.6477,929.5046) (0.6666,930.8302) (0.6784,928.2361) (0.7728,1106.8799) (0.7846,1094.1739) (0.7893,938.9986) (0.8294,941.7694) (0.8530,1108.1598) (0.8836,1101.5260) (0.9332,1269.9531) (0.9355,1103.8665) (0.9403,1276.6542) (1.0087,1266.4501) (1.1031,1288.4514) (1.1337,1295.1299) (1.1668,1451.5505) (1.1738,1458.3690) (1.2045,1295.2540) (1.2069,1457.7836) (1.2187,1645.9527) (1.2187,1292.8186) (1.2965,1468.3289) (1.3697,1483.6124) (1.3815,1634.2241) (1.4758,1647.3530) (1.5018,1814.9498) (1.5065,1632.1022) (1.5277,1636.6878) (1.5348,1473.2368) (1.5419,1641.8623) (1.5867,1485.4668) (1.5914,1476.5619) (1.6339,1809.9701) (1.6457,1650.0424) (1.6504,1805.7300) (1.6882,1828.0148) (1.7212,1826.4291) (1.7755,1823.7415) (1.7873,1836.7892) (1.8132,1676.6671) (1.8179,1847.1711) (1.8392,2001.0196) (1.8533,1655.1044) (1.9500,2009.4881) (2.0421,2029.2335) (2.0468,2032.7142) (2.0845,2006.5740) (2.0892,1840.9467) (2.1883,2033.9060) (2.2402,2183.9246) (2.2497,2026.8737) (2.2898,2030.5147) (2.2992,2022.5098) (2.3110,2179.6717) (2.3228,2192.7418) (2.3440,2033.2367) (2.3724,2197.8988) (2.3865,2210.4677) (2.3889,2048.1003) (2.3983,2197.4841) (2.4266,2200.2415) (2.4266,2201.9251) (2.4290,2208.2370) (2.4431,2371.5659) (2.4502,2195.9467) (2.4667,2206.0617) (2.5304,2378.4018) (2.5493,2212.4027) (2.5540,2212.2375) (2.6106,2379.2206) (2.6838,2383.9679) (2.7097,2386.7716) (2.7522,2384.1132) (2.7711,2393.6651) (2.8065,2395.1069) (2.8183,2219.9514) (2.8324,2394.9409) (2.8348,2394.6544) (2.8348,2235.2971) (2.8466,2391.9796) (2.8513,2396.8999) (2.8678,2397.7661) (2.8867,2397.9462) (2.8985,2396.0527) (2.9174,2395.2739) (2.9433,2401.3412) (2.9457,2403.2156) (2.9480,2400.2168) (2.9598,2399.5502) (2.9716,2397.2903) (2.9834,2405.1367) (2.9834,2404.6048) (3.0306,2406.0411) (3.0448,2407.2642)
};
\addlegendentry{Random (Baseline)}

\end{axis}
\end{tikzpicture}
}
\caption{Comparison of architectures discovered by our multi-objective search (Pareto Path) versus a uniform random sampling baseline. The horizontal coordinate is computational cost in GMACs, and the vertical coordinate is the structural fitness score of Eq.~\ref{eq:fitness_func}. Our search method consistently discovers an approximate Pareto-optimal frontier of architectures with significantly higher network fitness scores for a given computational budget. This pre-computed cache of elite subnets forms the basis of our path-guided training, avoiding the suboptimal, low-fitness architectures frequently chosen by the baseline's random sampler.}
\label{fig:search_comparison}
\end{figure}

We formally define the search space $\mathbb{S}$. For a supernet with $S$ stages, we define three distinct sets of discrete choices:
\begin{itemize}
    \item $D = \{d_{1}, \dots, d_{M_D}\}$: The set of choices for the number of blocks (depth) in a stage.
    \item $\mathcal{W} = \{w_{1}, \dots, w_{M_W}\}$: The set of choices for channel width multipliers.
    \item $E = \{e_{1}, \dots, e_{M_E}\}$: The set of choices for bottleneck expansion ratios.
\end{itemize}
Here, $M_D$, $M_W$, and $M_E$ denote the respective numbers of available choices.

An architecture $\mathcal{A} \in \mathbb{S}$ is uniquely represented by a genome consisting of three concatenated vectors:
\begin{equation}
\label{eq:arch_encoding}
\mathcal{A} = (\mathbf{d}, \mathbf{e}, \mathbf{w})
\end{equation}
where:
\begin{itemize}
    \item $\mathbf{d} \in D^S$: A depth vector of length $S$, where $d_i$ is the number of extra blocks in stage $i$.
    \item $\mathbf{e} \in E^{S \times N_{\text{blocks}}}$: An expansion-ratio vector, where $N_{\text{blocks}}$ is the maximum number of block slots per stage. A value is specified for every slot so that the genome has fixed length even when some slots are inactive.
    \item $\mathbf{w} \in \mathcal{W}^{S+1}$: A width vector of length $S+1$, specifying the multiplier for the stem ($w_0$) and each stage ($w_1, \dots, w_S$).
\end{itemize}

The full search space $\mathbb{S}$ is the Cartesian product of these vector spaces:
\begin{equation}
\label{eq:search_space}
\mathbb{S} = D^S \times E^{S \times N_{\text{blocks}}} \times \mathcal{W}^{S+1}
\end{equation}
This formal definition accurately reflects our implementation, where the genome is a flat vector concatenating depth, expansion, and width decisions. As demonstrated in our experimental setup (see Table~\ref{tab:supernet_capabilities}), this design enables a significantly broader range of model complexities compared to the baseline.

\subsection{An Entropy-Based, Multi-Objective Fitness Function}
\label{subsec:fitness_function}

To overcome random sampling, we require a mechanism to rank an architecture without performing costly training or validation for that candidate. We therefore develop a multi-objective structural fitness function, $\mathcal{F}(\mathcal{A})$, inspired by the mathematical design framework of DeepMAD \cite{Shen2023CVPR}, which balances network expressiveness (entropy) against architectural stability (effectiveness). We synthesize these concepts with structural guidelines (depth uniformity, channel monotonicity), reformulating them as penalty terms. Unlike the empirical loss in Eq.~\ref{eq:fl_objective}, $\mathcal{F}(\mathcal{A})$ depends only on the architecture and its derived structural and resource quantities; it does not use client data, supernet weights, or gradient updates.

Our unified fitness function for an architecture $\mathcal{A} \in \mathbb{S}$ is:

\begin{equation}
\label{eq:fitness_func}
\begin{aligned}
\mathcal{F}(\mathcal{A}) =  \sum_{j=1}^{S} \alpha_j H_j(\mathcal{A}) - \omega Q(\mathcal{A})
+ \lambda \rho(\mathcal{A}) - \gamma V(\mathcal{A})\\
\text{s.t.} \quad \rho(\mathcal{A}) \le \rho_0
\end{aligned}
\end{equation}

where $\alpha_j, \omega, \lambda, \gamma$ are weighting hyper-parameters and $\rho_0$ is the hard upper bound on effectiveness. Given the architecture genome $\mathcal{A} = (\mathbf{d}, \mathbf{e}, \mathbf{w})$, the components are defined as follows:

\begin{itemize}
    \item \textbf{Stage-wise Network Entropy ($H_j(\mathcal{A})$):} A proxy for the expressive power of stage $j$. Let $r_j$ be the feature map resolution at stage $j$'s end, and $c_{\text{out}, j}$ be the output width. Let $c_{\text{in}, \ell}(\mathcal{A})$ be the input width of layer $\ell$ (determined by $\mathbf{w}$ and $\mathbf{e}$), and $k_\ell$ be its kernel size. The entropy for stage $j$ sums the contributions of its active layers:
    \begin{equation}
    H_j(\mathcal{A}) = \log(r_j^2 \cdot c_{\text{out}, j}) \sum_{\ell \in \text{Stage}_j} \log(c_{\text{in}, \ell} \cdot k_\ell^2)
    \label{eq:entropy}
    \end{equation}
    Here, the inner summation iterates only over the active layers $\ell$ configured by the depth vector $\mathbf{d}$ for stage $j$. The logarithms summarize the architecture's structural scale. The terms $r_j^2c_{\text{out},j}$ and $c_{\text{in},\ell}k_\ell^2$ respectively reflect the size of the stage output representation and the per-layer convolutional connectivity. Applying logarithms converts multiplicative changes within these terms into additive contributions and prevents a single very wide layer from dominating the score. Thus, $H_j$ is an architecture-only structural score, not an empirical entropy computed from learned activations or labels. This use of structural quantities follows the information measure adapted from DeepMAD \cite{Shen2023CVPR}.
    
    \item \textbf{Effectiveness ($\rho(\mathcal{A})$):} A measure of architectural stability. Let $L(\mathcal{A})$ be the total depth of active layers.
    \begin{equation}
    \rho(\mathcal{A}) = \frac{L(\mathcal{A})}{\exp\left( \frac{1}{L(\mathcal{A})} \sum_{\ell \in \text{Active}} \log(c_{\text{in}, \ell}(\mathcal{A}) \cdot k_\ell^2) \right)}
    \end{equation}

    \item \textbf{Depth Uniformity Penalty ($Q(\mathcal{A})$):} A penalty for non-uniform stage depths, captured by the variance of the stage depth vector $\mathbf{d}$.
    \begin{equation}
    Q(\mathcal{A}) = \exp(\Var(\mathbf{d}))
    \end{equation}
    
    \item \textbf{Channel Monotonicity Penalty ($V(\mathcal{A})$):} A penalty for violating non-decreasing channel counts. Let $w_{\text{out}, i}(\mathcal{A})$ be the output channel count of stage $i$ derived from $\mathbf{w}$.
    \begin{equation}
    V(\mathcal{A}) = \sum_{i=0}^{S-1} \max(0, w_{\text{out}, i}(\mathcal{A}) - w_{\text{out}, i+1}(\mathcal{A}))
    \end{equation}
\end{itemize}
This unified function is maximized during offline cache construction and post-training deployment search, subject to hard constraints $\text{MACs}(\mathcal{A}) \le \text{Budget}$ and $\rho(\mathcal{A}) \le \rho_0$. Thus, empirical loss is used to train $W$, whereas structural fitness is used to rank candidate architectures. Including an empirical loss term in $\mathcal{F}$ would require executing and evaluating every candidate subnet, which is precisely the predictor-data collection cost that the structural search is designed to avoid.

\subsection{Search for Optimal Path Generation}
\label{subsec:offline_search}

We perform an extensive, offline search to generate a cache of elite subnets, $\mathcal{C}$, termed the ``optimal path cache." We discretize the computational range into $N$ target budgets, $\mathcal{B} = \{B_1, \dots, B_N\}$. We then solve $N$ independent optimization problems to find, for each budget $B_i \in \mathcal{B}$, the optimal architecture $\mathcal{A}^*_i$:
\begin{equation}
\label{eq:ga_objective}
\begin{aligned}
\mathcal{A}^*_i &= \underset{\mathcal{A} \in \mathbb{S}}{\arg\max} \; \mathcal{F}(\mathcal{A}) \\
&\text{subject to } \text{MACs}(\mathcal{A}) \le B_i, \; \rho(\mathcal{A}) \le \rho_0
\end{aligned}
\end{equation}
Each optimization in Eq.~\ref{eq:ga_objective} searches over architecture $\mathcal{A}$, rather than modifying supernet weights $W$, and is solved using a dedicated instance of a standard genetic algorithm. To ensure efficiency and validity, we employ a \textbf{rejection sampling initialization}: we randomly generate architectures and immediately discard any that violate the MACs or $\rho_0$ constraints, repeating until the initial population $P^{(0)}$ consists entirely of valid candidates.

The GA proceeds as follows:
\begin{enumerate}
    \item \textbf{Selection:} Tournament selection is used to choose parents from the current population $P^{(g)}$.
    \item \textbf{Crossover:} Single-point crossover combines the genomes $\mathcal{A}_a, \mathcal{A}_b$ to produce offspring.
    \item \textbf{Mutation:} With probability $p_m$, genes in $\mathbf{d}$, $\mathbf{e}$, or $\mathbf{w}$ are randomly reset to new valid values from their respective choice sets.
    \item \textbf{Elitism:} The best architecture found so far is preserved across generations.
\end{enumerate}
The final aggregated set of elite architectures, $\mathcal{C} = \{\mathcal{A}^*_1, \dots, \mathcal{A}^*_N\}$, forms a robust approximation of the true Pareto-optimal frontier. The size of this cache, $N$, is a hyperparameter chosen to ensure sufficient granularity across the FL computational spectrum; in our experiments, we analyze the sensitivity of the structural fitness statistics to $N$ (see Section~\ref{subsubsec:cache_analysis}).

\subsection{Federated Pareto Optimal Supernet Training}
\label{subsec:path_guided_training}

We formalize our training procedure in Algorithm~\ref{alg:deepfednas_training}. The central contribution is a fitness-generated Pareto path cache $\mathcal{C}$ and its use as a structured subnet-assignment curriculum, replacing the baseline's unguided random sampling. In each federated round $t$, the assignment of architecture $\mathcal{A}_k$ to client $k$ is governed by our Pareto Path assignment function $S(k, \mathcal{C})$. The algorithm retains the standard client-selection, local-update, and overlap-aware MaxNet aggregation mechanics of SuperFedNAS \cite{khare2023superfednas}:
\begin{equation}
\label{eq:path_sampling}
\mathcal{A}_k = S(k, \mathcal{C}) =
\begin{cases}
\mathcal{A}_{\min} & \text{if } k \in \mathcal{K}_{t, \min} \\
\mathcal{A}_{\max} & \text{if } k \in \mathcal{K}_{t, \max} \\
\text{Uniform}(\mathcal{C}) & \text{otherwise}
\end{cases}
\end{equation}
where $\mathcal{A}_{\min}, \mathcal{A}_{\max} \in \mathcal{C}$ are the boundary architectures from our cache, and $\mathcal{K}_{t, \min}, \mathcal{K}_{t, \max}$ are the sets of clients who have trained these boundary architectures least frequently. This strategy ensures the supernet's weights are consistently updated based on gradients from a curriculum of optimized, high-fitness architectures.

\begin{algorithm}[t]
\caption{DeepFedNAS: Federated Pareto Optimal Supernet Training}
\label{alg:deepfednas_training}
\begin{algorithmic}[1]
\REQUIRE Supernet weights $W_0$, clients $K$, rounds $T$, participation fraction $C$, learning rate $\eta$.
\ENSURE Optimized Supernet Weights $W_T$.

\STATE \textbf{Proposed server step:} Generate Pareto Path Cache $\mathcal{C}$ (Section~\ref{subsec:offline_search}).
\FOR{$t = 0$ to $T-1$}
    \STATE \textbf{Server:} Select subset of clients $\mathcal{K}_t$ (size $C \cdot K$).
    \FOR{each client $k \in \mathcal{K}_t$}
        \STATE \textbf{Proposed server step:} Determine architecture assignment $\mathcal{A}_k$ using Pareto Path Sampling:
        \STATE \hspace{1em} $\mathcal{A}_k \leftarrow S(k, \mathcal{C})$ \quad \textit{(Eq.~\ref{eq:path_sampling})}
        \STATE \textbf{Server:} Send active weights $w_{k,t} = \mathcal{G}(W_t, \mathcal{A}_k)$ to client $k$.
        \STATE \textbf{Client $k$:} Update weights locally on $\mathcal{D}_k$:
        \STATE \hspace{1em} $w'_{k,t} \leftarrow w_{k,t} - \eta \nabla \mathcal{L}_k(w_{k,t})$
        \STATE \textbf{Client $k$:} Compute update $\Delta_k = w'_{k,t} - w_{k,t}$ and upload.
    \ENDFOR
    \STATE \textbf{Server:} Aggregate using Overlap-Aware MaxNet:
    \FOR{each parameter $\theta$ in supernet}
        \STATE Compute $\Delta_\theta$ using binary masks $\mathbb{I}_k(\theta)$ \quad \textit{(Eq.~\ref{eq:maxnet_agg})}
        \STATE $W_{t+1}(\theta) \leftarrow W_t(\theta) + \eta_g \Delta_\theta$
    \ENDFOR
\ENDFOR
\RETURN $W_T$
\end{algorithmic}
\end{algorithm}

It is important to note that our Pareto Path Guided sampling introduces \emph{negligible overhead} to the per-round training cost. The Pareto cache $\mathcal{C}$ is pre-computed once offline (in $\sim$20 minutes on a standard CPU), and the per-round architecture assignment $S(k, \mathcal{C})$ is a simple lookup operation.

\subsection{Entropy-Based Fitness as a Predictor for Efficient Search}
\label{subsec:predictor_free}

A major contribution of DeepFedNAS is a Predictor-Free Search Method that circumvents the high computational cost of constructing learned accuracy predictors. This capability stems directly from our unified multi-objective fitness formulation, $\mathcal{F}(\mathcal{A})$, and our Path-Guided Training strategy.

We posit that by constraining the supernet training to a curriculum of high-fitness architectures (via the Pareto Path Subnet Cache), we condition the shared weights $W^*$ such that the structural fitness $\mathcal{F}(\mathcal{A})$ becomes a useful structural proxy for validation accuracy. While grounded in the information-theoretic concepts of entropy and effectiveness \cite{Shen2023CVPR}, our unified formulation transforms these theoretical bounds into a practical, searchable metric for federated supernets.

Formally, this relationship allows us to substitute the expensive expectation of accuracy with our low-cost architecture-based fitness function:
\begin{equation}
\label{eq:rank_correlation}
\underset{\mathcal{A} \in \mathbb{S}}{\arg\max} \; \E[\text{Acc}(\mathcal{A}, W^*)] \approx \underset{\mathcal{A} \in \mathbb{S}}{\arg\max} \; \mathcal{F}(\mathcal{A})
\end{equation}
Consequently, the deployment search for $\mathcal{A}^*_{\text{deploy}}$ becomes a direct optimization of Eq.~\ref{eq:fitness_func} using a genetic algorithm, completely obviating the need for a data-driven predictor pipeline. This allows the search to complete in seconds rather than hours. We provide empirical validation of the correlation between $\mathcal{F}(\mathcal{A})$ and true accuracy in Section~\ref{subsubsec:correlation_analysis}.

\subsection{Additional Hardware Constraints for Deployment}
\label{subsec:hardware_extensions}

Our framework supports hardware-aware deployment by integrating constraints such as parameters and latency into the final search.

\subsubsection{Integrating Model Parameters}
For memory-limited devices, we integrate an explicit upper bound on parameters ($\text{Params}(\mathcal{A})$) as a \emph{hard constraint} during the deployment search:
\begin{equation}
\label{eq:param_constraint}
\text{Params}(\mathcal{A}) \le \text{ParamBudget}_{\text{M}}
\end{equation}

\subsubsection{Integrating Inference Latency}
\label{subsubsec:inference_latency}
We employ a \emph{Latency Predictor Model} (LPM) to estimate inference latency, denoted as $\mathcal{L}_{\text{pred}}(\mathcal{A}, \text{Device})$. It is crucial to distinguish the computational nature of this LPM from the accuracy predictors used in prior work. Accuracy predictors require the generation of a dataset containing (subnet architecture, accuracy) pairs. Constructing this dataset is prohibitively expensive because measuring the ground-truth accuracy of a single subnet necessitates a full inference pass over the entire validation set. Iterating this process for thousands of architectures to train a predictor creates a computational bottleneck.

In contrast, training an LPM requires a dataset of (subnet architecture, latency) pairs. For each sampled architecture, one input sample (e.g., one image tensor) is profiled on the target device over a small number of repeated timed runs, and the measured inference times are averaged to reduce timing noise. This single-sample-per-architecture profiling protocol is independent of the validation dataset size and is substantially less costly than collecting an accuracy label, which requires a full validation-set evaluation. Thus, DeepFedNAS eliminates the costly accuracy-predictor pipeline while using a lightweight latency predictor only when latency is the deployment constraint.

The LPM is implemented as a lightweight MLP, $f_{\theta_{LPM}}$, trained offline. We formalize its components:
\begin{enumerate}
    \item \textbf{Architecture Featurization:} An architecture $\mathcal{A} = (\mathbf{d}, \mathbf{e}, \mathbf{w})$ is transformed into a fixed-length vector $v_{\mathcal{A}}$ via concatenation of values from the depth, expansion, and width vectors:
    \begin{equation}
    \begin{split}
    v_{\mathcal{A}} = \Concat(&\text{feat}(\mathbf{d}), \text{feat}(\mathbf{e}), \text{feat}(\mathbf{w}), \\
    &\text{MACs}(\mathcal{A}), \text{Params}(\mathcal{A}))
    \end{split}
    \end{equation}
    We explicitly include MACs and Parameters as features to enhance prediction accuracy.
    
    \item \textbf{Model Definition:} The LPM maps the feature vector to a scalar latency: $\mathcal{L}_{\text{pred}}(\mathcal{A}, \text{Device}) = f_{\theta_{LPM}}(v_{\mathcal{A}})$.
    
    \item \textbf{Training Objective:} The parameters $\theta_{LPM}$ are learned by minimizing the Mean Squared Error (MSE) on a measured dataset $\mathcal{D}_{\text{lat}} = \{(v_{\mathcal{A}_i}, \mathcal{L}_{\text{true}, i})\}_{i=1}^M$ for a specific target $\text{Device}$:
    \begin{equation}
    \underset{\theta_{LPM}}{\min} \; \frac{1}{M} \sum_{i=1}^M (f_{\theta_{LPM}}(v_{\mathcal{A}_i}) - \mathcal{L}_{\text{true}, i})^2
    \end{equation}
\end{enumerate}

This fast-to-train LPM allows latency to be incorporated into the deployment search as either a \textbf{hard constraint} ($\mathcal{L}_{\text{pred}}(\mathcal{A}, \text{Device}) \le \text{LatencyBudget}_{\text{ms}}$) or as a \textbf{soft objective} by modifying the fitness function:
\begin{equation}
\label{eq:latency_penalty}
\mathcal{F}_{\text{deploy}}(\mathcal{A}) = \mathcal{F}(\mathcal{A}) - \delta \cdot \mathcal{L}_{\text{pred}}(\mathcal{A}, \text{Device})
\end{equation}
where $\delta > 0$ is a tunable penalty weight.

The final, hardware-aware deployment search for $\mathcal{A}^*_{\text{deploy}}$ is formulated as a comprehensive, multi-objective optimization:
\begin{equation}
\label{eq:full_deployment_search}
\begin{aligned}
\mathcal{A}^*_{\text{deploy}}
    &= \underset{\mathcal{A} \in \mathbb{S}}{\arg\max}
     \; \mathcal{F}_{\text{deploy}}(\mathcal{A}) \\[0.5ex]
    &\text{subject to }
     \text{MACs}(\mathcal{A}) \le \text{Budget}_{\text{deploy}},\\
    &\phantom{\text{subject to }}
     \rho(\mathcal{A}) \le \rho_0,\\
    &\phantom{\text{subject to }}
     \text{Params}(\mathcal{A}) \le \text{ParamBudget}_{\text{M}},\\
    &\phantom{\text{subject to }}
     \mathcal{L}_{\text{pred}}(\mathcal{A},\text{Device})
     \le \text{LatencyBudget}_{\text{ms}}\\
    &\phantom{\text{subject to }}(\text{optional hard constraint})
\end{aligned}
\end{equation}
where $\mathcal{F}_{\text{deploy}}(\mathcal{A})$ uses Eq.~\ref{eq:latency_penalty} if a soft latency objective is chosen, or Eq.~\ref{eq:fitness_func} otherwise.

%% file: sections/04_experiments.tex
\section{Experiments}
\label{sec:experiments}

We conduct a series of experiments to validate the effectiveness of the DeepFedNAS framework. Our evaluation is designed to answer four primary research questions:
\begin{enumerate}
    \item \textbf{Does Pareto Optimal Supernet Training produce a superior supernet?} We compare subnets extracted from the DeepFedNAS-trained supernet against the SuperFedNAS baseline, which trains the same redesigned expanded supernet using random subnet sampling.
    \item \textbf{How effective is our Predictor-Free Search?} We evaluate the quality of architectures found by our on-demand fitness-driven search and quantify the reduction in computational cost compared to traditional accuracy-predictor pipelines.
    \item \textbf{How robust is our methodology?} We analyze the performance of our framework under varying FL conditions, including data heterogeneity and sparse client participation.
    \item \textbf{How does DeepFedNAS support hardware-aware deployment constraints?} We characterize the parameter and device-specific latency trade-offs of selected subnets and show how these quantities can be incorporated into the deployment search.
\end{enumerate}
We begin by detailing our experimental setup before presenting our main results and ablation studies.

\subsection{Experimental Setup}
\label{subsec:experimental_setup}

\begin{table}[b]
\centering
\caption{Architecture-space capability of the original SuperFedNAS search space and the redesigned expanded supernet.}
\label{tab:supernet_capabilities}
\begin{tabular}{lcc}
\toprule
\textbf{Metric} & \shortstack{\textbf{Original}\\\textbf{SuperFedNAS}} & \shortstack{\textbf{Redesigned}\\\textbf{expanded supernet}} \\
\midrule
Min. MACs (M) & 458.97 & \textbf{7.55} \\
Min. Params (M) & 10.40 & \textbf{0.13} \\
Max. MACs (M) & 3,403.37 & 3,403.37 \\
Max. Params (M) & 71.73 & 71.73 \\
\bottomrule
\end{tabular}
\end{table}

\begin{table*}[t]
\centering
\caption{Official-test comparison on image datasets at $\alpha=100$ and $C=0.4$. SuperFedNAS and DeepFedNAS use the same redesigned expanded supernet and the same four strict MAC intervals; they differ in their subnet-sampling strategy during training. Values are mean $\pm$ standard deviation across five post-training architecture searches using different search seeds.}
\label{tab:image_datasets}
\begin{tabular}{llccc}
\hline
\textbf{Billion MACs} & \textbf{Method} & \multicolumn{3}{c}{\textbf{Test Accuracy (\%)}} \\
\cline{3-5}
& & \textbf{CIFAR-10} & \textbf{CIFAR-100} & \textbf{CINIC-10} \\
\hline
\multirow{5}{*}{0.458-0.95}
& FedAvg\textsuperscript{\(\dagger\)} & $85.25 \pm 0.46$ & $43.19 \pm 0.54$ & $61.76 \pm 0.78$ \\
& FedNAS\textsuperscript{\(\dagger\)} & $77.33 \pm 0.31$ & $40.92 \pm 2.21$ & $58.15 \pm 0.18$ \\
& FedPNAS\textsuperscript{\(\dagger\)} & $88.83 \pm 0.50$ & $45.77 \pm 0.68$ & $64.30 \pm 0.98$ \\
& SuperFedNAS & $93.86 \pm 0.21$ & $71.21 \pm 0.41$ & $78.74 \pm 0.30$ \\
& \textbf{DeepFedNAS} & $\mathbf{94.43 \pm 0.23}$ & $\mathbf{73.09 \pm 0.41}$ & $\mathbf{81.81 \pm 0.38}$ \\
\hline
\multirow{4}{*}{0.95-1.45}
& FedAvg\textsuperscript{\(\dagger\)} & $86.36 \pm 0.22$ & $43.92 \pm 0.57$ & $63.00 \pm 0.17$ \\
& FedPNAS\textsuperscript{\(\dagger\)} & $89.27 \pm 0.51$ & $47.80 \pm 0.26$ & $66.74 \pm 0.32$ \\
& SuperFedNAS & $94.34 \pm 0.07$ & $72.13 \pm 0.17$ & $79.57 \pm 0.00$ \\
& \textbf{DeepFedNAS} & $\mathbf{94.86 \pm 0.08}$ & $\mathbf{73.76 \pm 0.29}$ & $\mathbf{83.16 \pm 0.12}$ \\
\hline
\multirow{4}{*}{1.45-2.45}
& FedAvg\textsuperscript{\(\dagger\)} & $87.59 \pm 0.27$ & $44.40 \pm 0.56$ & $64.00 \pm 0.07$ \\
& FedNAS\textsuperscript{\(\dagger\)} & $86.41 \pm 0.10$ & $55.82 \pm 0.29$ & $69.97 \pm 0.27$ \\
& SuperFedNAS & $94.29 \pm 0.28$ & $72.49 \pm 0.30$ & $79.73 \pm 0.63$ \\
& \textbf{DeepFedNAS} & $\mathbf{95.01 \pm 0.03}$ & $\mathbf{74.85 \pm 0.10}$ & $\mathbf{83.95 \pm 0.04}$ \\
\hline
\multirow{4}{*}{2.45-3.403}
& FedAvg\textsuperscript{\(\dagger\)} & $89.44 \pm 0.67$ & $45.00 \pm 0.27$ & $66.02 \pm 0.13$ \\
& FedNAS\textsuperscript{\(\dagger\)} & $89.43 \pm 0.36$ & $58.39 \pm 0.23$ & $71.93 \pm 0.13$ \\
& SuperFedNAS & $94.02 \pm 0.02$ & $72.36 \pm 0.10$ & $79.25 \pm 0.09$ \\
& \textbf{DeepFedNAS} & $\mathbf{95.06 \pm 0.00}$ & $\mathbf{74.96 \pm 0.00}$ & $\mathbf{84.05 \pm 0.00}$ \\
\hline
\multicolumn{5}{l}{\footnotesize\textsuperscript{\(\dagger\)} Published results from SuperFedNAS~\cite{khare2023superfednas}, provided for reference; the source uses 0.45--0.95 and 2.45--3.75 B for the endpoint MAC ranges.}\\
\end{tabular}
\end{table*}

\paragraph{Datasets and Data Partitioning}
Our experiments use \textbf{CIFAR-10}, \textbf{CIFAR-100}, and \textbf{CINIC-10}. For each CIFAR dataset, we create a fixed stratified 45,000/5,000 training/validation split from the official training set and partition only the 45,000 training images among clients. For CINIC-10, we use its official training, validation, and test directories for the corresponding roles. Statistical heterogeneity is introduced by partitioning training data among clients using a Dirichlet distribution \cite{FedDC}. The main comparison uses $\alpha=100$; the heterogeneity analysis uses $\alpha \in \{100,1,0.1\}$ (Section~\ref{subsec:noniid_results}). We use $K=20$ clients for CIFAR-10 and CIFAR-100 and $K=100$ clients for CINIC-10.

\paragraph{Evaluation Protocol} 
The validation split is used for checkpoint selection and for any data-dependent search decisions, while the DeepFedNAS deployment search itself ranks candidates using the architecture-based fitness $\mathcal{F}(\mathcal{A})$ without validation-set evaluation. The best checkpoint maximizes the mean validation accuracy of the four reference subnets evaluated every 20 rounds. For the SuperFedNAS baseline, validation data are also the only data used to construct and fit its accuracy predictor. We perform five post-training architecture searches with seeds 42--46, record the selected architectures, and then evaluate them once on the untouched official test set. The searches use a population of 256, 512 generations, and mutation probability 0.3. Thus, no test example is used for federated training, checkpoint selection, SuperFedNAS predictor construction, architecture search, correlation analysis, or architecture selection. Reported $\pm$ values summarize variation across the five searches. A value of $0.00$ means that all five searches selected the same executable architecture, which commonly occurs at the all-maximum endpoint.

\begin{table*}[t]
\centering
\caption{Official-test comparison across degrees of non-IID data on CIFAR-10 at $C=0.4$. SuperFedNAS and DeepFedNAS use the same redesigned expanded supernet and strict MAC intervals. Values are mean $\pm$ standard deviation across five post-training architecture searches using different search seeds.}
\label{tab:noniid}
\begin{tabular}{llccc}
\hline
\textbf{Billion MACs} & \textbf{Method} & \multicolumn{3}{c}{\textbf{Test Accuracy (\%)}} \\
\cline{3-5}
& & \textbf{non-iid-100 ($\alpha=100$)} & \textbf{non-iid-1 ($\alpha=1$)} & \textbf{non-iid-0.1 ($\alpha=0.1$)} \\
\hline
\multirow{5}{*}{0.458-0.95}
& FedAvg\textsuperscript{\(\dagger\)} & $85.25 \pm 0.46$ & $83.42 \pm 0.19$ & $77.15 \pm 2.50$ \\
& FedNAS\textsuperscript{\(\dagger\)} & $77.33 \pm 0.31$ & $71.38 \pm 0.37$ & $51.57 \pm 3.32$ \\
& FedPNAS\textsuperscript{\(\dagger\)} & $88.83 \pm 0.50$ & $85.70 \pm 0.40$ & $78.73 \pm 0.45$ \\
& SuperFedNAS & $93.86 \pm 0.21$ & $92.40 \pm 0.89$ & $71.17 \pm 1.69$ \\
& \textbf{DeepFedNAS} & $\mathbf{94.43 \pm 0.23}$ & $\mathbf{93.13 \pm 0.19}$ & $\mathbf{75.72 \pm 0.22}$ \\
\hline
\multirow{4}{*}{0.95-1.45}
& FedAvg\textsuperscript{\(\dagger\)} & $86.36 \pm 0.22$ & $84.65 \pm 0.11$ & $77.99 \pm 1.60$ \\
& FedPNAS\textsuperscript{\(\dagger\)} & $89.27 \pm 0.51$ & $87.53 \pm 0.32$ & $81.13 \pm 0.40$ \\
& SuperFedNAS & $94.34 \pm 0.07$ & $92.61 \pm 0.92$ & $72.85 \pm 0.36$ \\
& \textbf{DeepFedNAS} & $\mathbf{94.86 \pm 0.08}$ & $\mathbf{93.56 \pm 0.03}$ & $\mathbf{76.25 \pm 0.19}$ \\
\hline
\multirow{4}{*}{1.45-2.45}
& FedAvg\textsuperscript{\(\dagger\)} & $87.59 \pm 0.27$ & $86.14 \pm 0.23$ & $79.93 \pm 1.34$ \\
& FedNAS\textsuperscript{\(\dagger\)} & $86.41 \pm 0.10$ & $82.13 \pm 0.65$ & $65.03 \pm 2.57$ \\
& SuperFedNAS & $94.29 \pm 0.28$ & $93.10 \pm 0.80$ & $72.91 \pm 0.13$ \\
& \textbf{DeepFedNAS} & $\mathbf{95.01 \pm 0.03}$ & $\mathbf{93.91 \pm 0.03}$ & $\mathbf{76.91 \pm 0.08}$ \\
\hline
\multirow{4}{*}{2.45-3.403}
& FedAvg\textsuperscript{\(\dagger\)} & $89.44 \pm 0.67$ & $87.88 \pm 0.70$ & $81.24 \pm 1.99$ \\
& FedNAS\textsuperscript{\(\dagger\)} & $89.43 \pm 0.36$ & $85.85 \pm 0.35$ & $68.13 \pm 5.04$ \\
& SuperFedNAS & $94.02 \pm 0.02$ & $92.54 \pm 0.04$ & $72.87 \pm 0.20$ \\
& \textbf{DeepFedNAS} & $\mathbf{95.06 \pm 0.00}$ & $\mathbf{93.88 \pm 0.00}$ & $\mathbf{77.00 \pm 0.00}$ \\
\hline
\multicolumn{5}{l}{\footnotesize\textsuperscript{\(\dagger\)} Published results from SuperFedNAS~\cite{khare2023superfednas}, provided for reference; the source uses 0.45--0.95 and 2.45--3.75 B for the endpoint MAC ranges.}\\
\end{tabular}
\end{table*}

\paragraph{Supernetwork Architecture}
To evaluate our method across a complex design space, we utilize the generic ResNet-style supernet framework detailed in Section~\ref{subsec:supernet_design}. For these experiments, we instantiate a 4-stage supernet with base channel sizes of [256, 512, 1024, 2048]. The search space is defined by three variable dimensions: depth $d_i \in \{0,1,2\}$ additional blocks per stage, corresponding to $1+d_i \in \{1,2,3\}$ total blocks per stage; width $\mathbf{w}$ with multipliers from 0.1 to 1.0 in steps of 0.1; and bottleneck expansion ratios $\mathbf{e} \in \{0.1, 0.14, 0.18, 0.22, 0.25\}$. This results in a combinatorial space of approximately $1.98 \times 10^{15}$ architectures, significantly larger than prior work (see Table~\ref{tab:supernet_capabilities}).

\paragraph{Search Space Exploration and Budget Alignment}
The four predefined MAC intervals within 0.458 to 3.403 billion MACs define the controlled evaluation range used to compare DeepFedNAS with SuperFedNAS. Although the redesigned expanded supernet spans 7.55 M to approximately 3,403 M MACs, the DeepFedNAS Pareto-path cache is generated within the overall 0.458 to 3.403 billion-MAC range for supernet training. The four intervals are then used to report and compare the searched subnets. Both DeepFedNAS and SuperFedNAS use the same redesigned expanded supernet and evaluation intervals. Their training differs in the subnet-assignment policy: SuperFedNAS uses random sampling from the redesigned supernet, whereas DeepFedNAS draws assignments from the Pareto-path cache. Daggered results for preceding methods are published reference values.

\begin{table*}[t]
\centering
\caption{Official-test comparison across client participation rates on CIFAR-10 at $\alpha=100$. SuperFedNAS and DeepFedNAS use the same redesigned expanded supernet and strict MAC intervals. Values are mean $\pm$ standard deviation across five post-training architecture searches using different search seeds.}
\label{tab:participation}
\resizebox{\textwidth}{!}{%
\begin{tabular}{llcccc}
\hline
\textbf{Billion MACs} & \textbf{Method} & \multicolumn{4}{c}{\textbf{Test Accuracy (\%)}} \\
\cline{3-6}
& & \textbf{C = 0.1} & \textbf{C = 0.2} & \textbf{C = 0.4} & \textbf{C = 0.6} \\
\hline
\multirow{4}{*}{0.458-0.95}
& FedNAS\textsuperscript{\(\dagger\)} & -- & $76.23 \pm 0.50$ & $77.33 \pm 0.31$ & -- \\
& FedPNAS\textsuperscript{\(\dagger\)} & -- & $86.63 \pm 0.51$ & $88.83 \pm 0.50$ & -- \\
& SuperFedNAS & $91.70 \pm 0.64$ & $87.64 \pm 5.20$ & $93.86 \pm 0.21$ & $93.03 \pm 0.27$ \\
& \textbf{DeepFedNAS} & $\mathbf{92.41 \pm 0.35}$ & $\mathbf{94.08 \pm 0.09}$ & $\mathbf{94.43 \pm 0.23}$ & $\mathbf{94.72 \pm 0.19}$ \\
\hline
\multirow{3}{*}{0.95-1.45}
& FedPNAS\textsuperscript{\(\dagger\)} & -- & $87.83 \pm 0.21$ & $89.27 \pm 0.51$ & -- \\
& SuperFedNAS & $\mathbf{92.91 \pm 0.09}$ & $91.89 \pm 0.77$ & $94.34 \pm 0.07$ & $94.19 \pm 0.28$ \\
& \textbf{DeepFedNAS} & $92.67 \pm 0.10$ & $\mathbf{94.46 \pm 0.02}$ & $\mathbf{94.86 \pm 0.08}$ & $\mathbf{95.19 \pm 0.08}$ \\
\hline
\multirow{3}{*}{1.45-2.45}
& FedNAS\textsuperscript{\(\dagger\)} & -- & $84.65 \pm 0.14$ & $86.41 \pm 0.10$ & -- \\
& SuperFedNAS & $93.06 \pm 0.23$ & $93.14 \pm 0.35$ & $94.29 \pm 0.28$ & $93.80 \pm 0.01$ \\
& \textbf{DeepFedNAS} & $\mathbf{93.14 \pm 0.09}$ & $\mathbf{94.55 \pm 0.07}$ & $\mathbf{95.01 \pm 0.03}$ & $\mathbf{95.49 \pm 0.10}$ \\
\hline
\multirow{3}{*}{2.45-3.403}
& FedNAS\textsuperscript{\(\dagger\)} & -- & $88.00 \pm 0.38$ & $89.43 \pm 0.36$ & -- \\
& SuperFedNAS & $93.25 \pm 0.17$ & $92.61 \pm 0.01$ & $94.02 \pm 0.02$ & $94.12 \pm 0.01$ \\
& \textbf{DeepFedNAS} & $\mathbf{93.31 \pm 0.00}$ & $\mathbf{94.69 \pm 0.00}$ & $\mathbf{95.06 \pm 0.00}$ & $\mathbf{95.43 \pm 0.00}$ \\
\hline
\multicolumn{6}{l}{\footnotesize\textsuperscript{\(\dagger\)} Published results from SuperFedNAS~\cite{khare2023superfednas}, provided for reference; the source uses 0.45--0.95 and 2.45--3.75 B for the endpoint MAC ranges.}\\
\end{tabular}
}
\end{table*}

\paragraph{Pareto-Cache Construction}
For the DeepFedNAS experiments, the Pareto-path cache contains 60 subnets. The fitness function in Eq.~\ref{eq:fitness_func} is instantiated with $S=4$, $\alpha_j=1$ for $j=1,\ldots,4$, $\omega=10$, $\lambda=1000$, and $\gamma=1$. The hard effectiveness constraint uses $\rho_0=0.51$.

\paragraph{Federated Training}
For the controlled comparisons, both methods use the same client partitions, training seed, redesigned expanded supernet, data-augmentation configuration, and aggregation schedule. We use training seed 0, a training batch size of 64, and a validation batch size of 512. The augmentation combines RandAugment~\cite{cubuk2020randaugment} with alternating Mixup~\cite{zhang2018mixup} and CutMix~\cite{yun2019cutmix}. Local updates use SGD with initial learning rate 0.1, weight decay $5\times10^{-4}$, cosine learning-rate decay, and a gradient-clipping threshold of 10. We use the dynamic weighted-averaging schedule of SuperFedNAS \cite{khare2023superfednas}, which anneals the aggregation weights from an emphasis on the largest subnet to a uniform distribution over the first 80\% of training rounds. Each round performs five local epochs. CIFAR-10 and CINIC-10 are trained for 1,500 communication rounds, and CIFAR-100 for 2,000 rounds. The main comparison uses $C=0.4$; the participation study varies $C$ as specified in Table~\ref{tab:participation}.

\paragraph{Baselines for Comparison}
Our primary controlled baseline, \textbf{SuperFedNAS}, uses the same redesigned expanded supernet and operating intervals as DeepFedNAS but assigns subnets by random sampling rather than Pareto-path sampling. This keeps the supernet design and evaluation protocol the same while directly comparing the two subnet-assignment policies. Published FedAvg, FedNAS, and FedPNAS results from the SuperFedNAS paper are also included for reference and marked with a dagger in the tables.

\subsection{Comparison on Image Datasets}
\label{subsec:image_results}

We first evaluate DeepFedNAS on three benchmark datasets. Table~\ref{tab:image_datasets} shows that DeepFedNAS exceeds SuperFedNAS in every reported interval. The equal-interval mean advantage is 0.71 percentage points on CIFAR-10, 2.12 points on CIFAR-100, and 3.92 points on CINIC-10. The results validate our primary hypothesis: training the supernet on a curriculum of elite, high-fitness architectures yields shared weights that are better conditioned for fine-grained architectural search.

\subsection{Impact of Data Heterogeneity}
\label{subsec:noniid_results}

We evaluate robustness to data heterogeneity by varying the Dirichlet parameter $\alpha \in \{100, 1, 0.1\}$ \cite{hsu2019measuringeffectsnonidenticaldata}. As shown in Table~\ref{tab:noniid}, the advantage of DeepFedNAS over SuperFedNAS is 0.73--1.34 points at $\alpha=1$ and 3.40--4.55 points at $\alpha=0.1$. The results suggest that Pareto-path subnet assignment remains beneficial as client data become more heterogeneous. The widening advantage at lower $\alpha$ suggests that Pareto-path subnet assignment is more robust to heterogeneous client updates.

\subsection{Impact of Client Participation}
\label{subsec:participation_results}

We vary the client participation rate as $C \in \{0.1, 0.2, 0.4, 0.6\}$. Table~\ref{tab:participation} presents the CIFAR-10 results. DeepFedNAS leads in all four intervals at $C=0.6$ and $C=0.2$, with average advantages of 1.42 and 3.12 points, respectively. These results indicate that Pareto-guided training can maintain strong subnet accuracy with fewer participating clients. In particular, DeepFedNAS at $C=0.2$ exceeds SuperFedNAS at $C=0.4$ in all four MAC intervals, despite using half as many clients per round. This suggests a potential reduction in communication requirements without sacrificing the accuracy of the selected subnets. At $C=0.1$, only two clients participate in each round, and both methods train the same largest and smallest subnets. No intermediate subnet is assigned in this setting. Accordingly, the methods are near parity: DeepFedNAS has a mean advantage of 0.15 points, while SuperFedNAS is higher by 0.24 points in the 0.95--1.45 B interval.

\subsection{Predictor-Free Search Efficiency}
\label{subsec:search_efficiency}

A primary contribution of this work is the elimination of the costly post-training accuracy-predictor pipeline. We therefore report both its hardware-independent evaluation workload and measured wall-clock cost in Table~\ref{tab:search_cost}.

In our CIFAR-10 benchmark, the baseline SuperFedNAS approach is burdened by a substantial post-training setup cost: constructing its accuracy predictor requires evaluating 10,000 architectures over the complete 5,000-image validation split. This entails 10,000 full validation sweeps, corresponding to 50 million image-level forward evaluations. DeepFedNAS eliminates this accuracy-label collection stage altogether. On an NVIDIA RTX A5000, the baseline predictor workflow takes approximately 3.9 hours, whereas DeepFedNAS directly identifies an architecture optimized for the target resource budget in $\sim$20 seconds on a CPU. DeepFedNAS performs a one-time, approximately 20-minute CPU-based Pareto-cache construction step before supernet training.

\begin{table*}[t]
\centering
\caption{Auxiliary architecture-selection costs for the CIFAR-10 benchmark. Constructing the SuperFedNAS accuracy predictor requires evaluating 10,000 subnets over the 5,000-image validation split, totaling 50 million image-level forward evaluations. DeepFedNAS's Predictor-Free Search eliminates this evaluation workload. Measured execution times are reported together with the corresponding hardware.}
\label{tab:search_cost}
\begin{tabular}{lcc}
\hline
\textbf{Pipeline component} & \textbf{SuperFedNAS (Baseline)} & \textbf{DeepFedNAS (Ours)} \\
\hline
\multicolumn{3}{l}{\textbf{Before supernet training}} \\
\qquad Pareto-cache generation & Not required & $\sim$20 min on CPU \\
\multicolumn{3}{l}{\textbf{After supernet training}} \\
\qquad Subnets evaluated for accuracy-predictor labels & 10,000 & \textbf{0} \\
\qquad Image-level forward evaluations & 50 million & \textbf{0} \\
\qquad Accuracy-predictor construction & $\sim$3.9 h on RTX A5000 & \textbf{Not required} \\
\qquad Constrained search per target & $\sim$21.7 s on CPU & $\sim$20 s on CPU \\
\hline
\textbf{Total auxiliary setup and first search} & \textbf{$\sim$3.91 h} & \textbf{$\sim$20.33 min} \\
\hline
\end{tabular}
\end{table*}

\subsection{Parameter and Communication Efficiency}
Parameter count is a useful proxy for the payload of a deployed model, although actual communication traffic also depends on the federated protocol and model representation. Figure~\ref{fig:combined_accuracy_params} reports the official-test accuracy and parameter counts of the selected subnets within the four predefined MAC intervals: 0.458--0.95B, 0.95--1.45B, 1.45--2.45B, and 2.45--3.403B.

On CIFAR-100, DeepFedNAS attains 73.09\% mean test accuracy with 18.88M parameters in the smallest interval. The best mean accuracy attained by SuperFedNAS is 72.49\% with 55.70M parameters. Thus, DeepFedNAS's smallest-interval operating point exceeds SuperFedNAS's best mean accuracy while using $2.95\times$ fewer parameters. DeepFedNAS also uses fewer mean parameters in the first three paired intervals while attaining higher accuracy in all four. At the largest interval, it selects the all-maximum endpoint and therefore uses more parameters than SuperFedNAS. On CINIC-10, DeepFedNAS has higher accuracy in all four intervals but uses more parameters, illustrating that the controlled benefit is improved subnet accuracy under the matched MAC budgets rather than a universal reduction in parameter count.

\begin{figure}[ht]
\centering
\begin{subfigure}{0.85\columnwidth}
\centering
\begin{tikzpicture}
 \begin{axis}[
 width=0.85\columnwidth, 
 height=4.5cm, 
 xlabel={Parameters (M)},
 ylabel={Accuracy (\%)},
 label style={font=\scriptsize}, 
 tick label style={font=\scriptsize}, 
 legend style={at={(0.5,-0.3)},anchor=north,legend columns=2,font=\scriptsize}, 
 grid=major,
 xmin=15, xmax=75,
 ymin=70.5, ymax=75.5,
 major tick length=2pt, 
 minor tick num=1 
 ]
 \addplot[red, mark=o, thick] table[row sep=crcr]{
  24.55 71.21 \\
  41.08 72.13 \\
  55.70 72.49 \\
  64.79 72.36 \\
 };
 \addlegendentry{SuperFedNAS}
 \addplot[blue, mark=square, thick] table[row sep=crcr]{
  18.88 73.09 \\
  28.04 73.76 \\
  46.46 74.85 \\
  71.91 74.96 \\
 };
 \addlegendentry{DeepFedNAS}
 \node[anchor=north west, font=\tiny] at (axis cs:24.55, 71.21) {0.93};
 \node[anchor=north west, font=\tiny] at (axis cs:41.08, 72.13) {1.41};
 \node[anchor=north west, font=\tiny] at (axis cs:55.70, 72.49) {2.42};
 \node[anchor=north west, font=\tiny] at (axis cs:64.79, 72.36) {2.94};
 \node[anchor=south west, font=\tiny] at (axis cs:18.88, 73.09) {0.95};
 \node[anchor=south east, font=\tiny] at (axis cs:28.04, 73.76) {1.45};
 \node[anchor=south east, font=\tiny] at (axis cs:46.46, 74.85) {2.44};
 \node[anchor=south east, font=\tiny] at (axis cs:71.91, 74.96) {3.40};
 \end{axis}
\end{tikzpicture}
\caption{CIFAR-100}
\label{fig:cifar100_accuracy_params}
\end{subfigure}

\vspace{0.2cm} 

\begin{subfigure}{0.85\columnwidth}
\centering
\begin{tikzpicture}
 \begin{axis}[
 width=0.85\columnwidth, 
 height=4.5cm, 
 xlabel={Parameters (M)},
 ylabel={Accuracy (\%)},
 label style={font=\scriptsize}, 
 tick label style={font=\scriptsize}, 
 legend style={at={(0.5,-0.3)},anchor=north,legend columns=2,font=\scriptsize}, 
 grid=major,
 xmin=0, xmax=75,
 ymin=78, ymax=85,
 major tick length=2pt, 
 minor tick num=1 
 ]
 \addplot[red, mark=o, thick] table[row sep=crcr]{
  11.87 78.74 \\
  14.42 79.57 \\
  15.87 79.73 \\
  46.17 79.25 \\
 };
 \addlegendentry{SuperFedNAS}
 \addplot[blue, mark=square, thick] table[row sep=crcr]{
  18.57 81.81 \\
  26.50 83.16 \\
  46.28 83.95 \\
  71.73 84.05 \\
 };
 \addlegendentry{DeepFedNAS}
 \node[anchor=north west, font=\tiny] at (axis cs:11.87, 78.74) {0.85};
 \node[anchor=south east, font=\tiny] at (axis cs:14.42, 79.57) {1.36};
 \node[anchor=north west, font=\tiny] at (axis cs:15.87, 79.73) {1.87};
 \node[anchor=north west, font=\tiny] at (axis cs:46.17, 79.25) {2.57};
 \node[anchor=south east, font=\tiny] at (axis cs:18.57, 81.81) {0.95};
 \node[anchor=south east, font=\tiny] at (axis cs:26.50, 83.16) {1.45};
 \node[anchor=south east, font=\tiny] at (axis cs:46.28, 83.95) {2.44};
 \node[anchor=south east, font=\tiny] at (axis cs:71.73, 84.05) {3.40};
 \end{axis}
\end{tikzpicture}
\caption{CINIC-10}
\label{fig:cinic10_accuracy_params}
\end{subfigure}

\caption{
    \textbf{Accuracy vs. Parameters (Millions).}
    Official-test accuracy against model parameters (millions) on (a) CIFAR-100 and (b) CINIC-10. DeepFedNAS subnets are blue squares; SuperFedNAS subnets are red circles. Each point is the mean result from five post-training searches within one predefined MAC interval (0.458--0.95B, 0.95--1.45B, 1.45--2.45B, or 2.45--3.403B); annotations denote mean actual MACs in billions.
}
\label{fig:combined_accuracy_params}
\end{figure}

\subsection{Hardware-Aware Latency Characterization}

Neural-network deployment requires balancing accuracy with device-specific
inference latency. DeepFedNAS can incorporate a latency constraint or objective
in its deployment search. We measured the four CIFAR-10 DeepFedNAS operating
points selected within the predefined MAC intervals on an Intel Xeon Silver
4210R CPU (2.40~GHz) and an NVIDIA RTX A5000 GPU. Figure~\ref{fig:latency_accuracy_tradeoff_cpu_gpu} shows the resulting latency--accuracy sweep. On the 16-thread CPU, mean single-input inference latency increases from 16.80 to 38.74~ms as mean actual MACs increase from 0.95B to 3.40B, while mean official-test accuracy increases from 94.43\% to 95.06\%. On the A5000, single-input inference latency is comparatively flat at 3.82--4.17~ms over the same operating points. These device-specific responses show why MAC count alone is insufficient to characterize deployment latency.

\begin{figure}[h!] 
    \centering
    \begin{tikzpicture}
    \begin{axis}[
        title={Official-Test Accuracy vs. Single-Input Latency},
        xlabel={Single-Input Latency (ms)},
        ylabel={Official-Test Accuracy (\%)},
        xmin=2.5, xmax=42,
        ymin=94.2, ymax=95.2,
        grid=major,
        legend pos=south east,
    ]

    \addplot[only marks, mark=*, blue] coordinates {
        (16.795, 94.428)
        (19.320, 94.862)
        (24.422, 95.008)
        (38.741, 95.060)
    };
    \addlegendentry{CPU}

    \addplot[only marks, mark=triangle*, red] coordinates {
        (3.824, 94.428)
        (3.890, 94.862)
        (3.857, 95.008)
        (4.170, 95.060)
    };
    \addlegendentry{A5000 GPU}

    \node[anchor=south east, font=\tiny] at (axis cs:16.795, 94.428) {0.95};
    \node[anchor=south east, font=\tiny] at (axis cs:19.320, 94.862) {1.45};
    \node[anchor=south east, font=\tiny] at (axis cs:24.422, 95.008) {2.44};
    \node[anchor=south east, font=\tiny] at (axis cs:38.741, 95.060) {3.40};

    \node[anchor=south west, font=\tiny] at (axis cs:3.824, 94.428) {0.95};
    \node[anchor=north west, font=\tiny] at (axis cs:3.890, 94.862) {1.45};
    \node[anchor=south east, font=\tiny] at (axis cs:3.857, 95.008) {2.44};
    \node[anchor=south west, font=\tiny] at (axis cs:4.170, 95.060) {3.40};

    \end{axis}
    \end{tikzpicture}
    \caption{
        \textbf{Accuracy vs. Single-Input Latency.} Official-test DeepFedNAS operating points on CIFAR-10. Circles denote the Intel Xeon Silver 4210R CPU (2.40~GHz; 16 PyTorch threads), and triangles the NVIDIA RTX A5000 GPU. Each point averages five post-training searches within one predefined MAC interval. Single-input inference latency is measured using one float32 $(1,3,32,32)$ input per architecture, profiled over a small number of repeated timed runs after a brief warm-up. Annotations show mean actual MACs (in billions).
    }
    \label{fig:latency_accuracy_tradeoff_cpu_gpu}
\end{figure}

\subsection{Implications for IoT Deployment}
\label{subsec:iot_deployment}

The efficiency gains demonstrated by DeepFedNAS carry direct practical implications for deploying federated learning across heterogeneous IoT device fleets. We analyze three key dimensions: communication overhead, fleet-wide architecture specialization, and scalability to new device classes.

\paragraph{Communication Overhead Reduction}
In federated IoT systems, the volume of model parameters transmitted per round affects bandwidth consumption and communication energy. As shown in Fig.~\ref{fig:combined_accuracy_params}, the controlled CIFAR-100 comparison finds that DeepFedNAS uses fewer mean parameters in the first three matched MAC intervals while attaining higher accuracy in every interval. This provides a useful model-payload proxy for deployments that select subnets in these intervals. Actual network traffic additionally depends on participation, serialization, protocol overhead, and any compression, and is therefore not estimated from parameter counts alone.

\paragraph{Fleet-Wide Architecture Specialization}
A defining characteristic of IoT federations is the coexistence of devices with vastly different resource budgets. Consider a representative smart-home deployment comprising: (i)~a security camera with an edge GPU ($\sim$2--3B MACs budget), (ii)~a smart thermostat with a mid-range processor ($\sim$1B MACs), and (iii)~a lightweight sensor hub ($\sim$0.5B MACs). After a single supernet training phase, DeepFedNAS can specialize an architecture for each device class through independent CPU searches of $\sim$20 seconds each. The SuperFedNAS baseline, by contrast, first requires construction of its 10,000-architecture accuracy predictor, which takes approximately 3.9 hours on an NVIDIA RTX A5000. DeepFedNAS instead uses a one-time, approximately 20-minute CPU Pareto-cache construction step before supernet training, allowing a single trained supernet to serve diverse deployment budgets without post-training accuracy-label collection or accuracy-predictor fitting.

\paragraph{Scalability to New Device Classes}
IoT device ecosystems are not static; new hardware platforms, firmware revisions, and deployment contexts emerge continuously. A critical advantage of the DeepFedNAS pipeline is that accommodating a new device class requires \emph{only} a lightweight search against the pre-trained supernet. No retraining, new accuracy-label collection, or accuracy-predictor fitting is needed. If a latency constraint is desired, only lightweight single-sample-per-architecture profiling (Section~\ref{subsubsec:inference_latency}) is required to collect labels for the latency predictor, a process that is substantially cheaper than the full accuracy evaluation pipeline. This ``plug-and-search'' capability is essential for IoT operators managing evolving, heterogeneous fleets at scale.

\subsection{Ablation Studies and Analysis}
\label{subsec:ablation}

In this section, we empirically validate the robustness of the DeepFedNAS framework, specifically analyzing the sensitivity of the cache size, the individual contributions of our multi-objective fitness components, and the efficacy of the fitness function as a predictor-free accuracy proxy.

\subsubsection{Sensitivity Analysis of Subnet Cache Size}
\label{subsubsec:cache_analysis}

The size of the optimal path cache, $N$, determines the granularity of the curriculum used during Federated Supernet Training. We conducted a structural sensitivity analysis using independently generated caches with equidistant sampling across the computational budget. We measured the mean and standard deviation of structural fitness as $N$ increased from 4 to 100 in increments of 4.

\begin{figure}[h]
\centering
\begin{subfigure}[t]{0.48\columnwidth} 
\centering
\begin{tikzpicture}
  \begin{axis}[
    width=\linewidth, 
    height=4.0cm, 
    xlabel={Cache Size ($N$)},
    ylabel={Avg. Fitness},
    label style={font=\scriptsize}, 
    tick label style={font=\scriptsize}, 
    grid=major,
    ymin=2150, ymax=2350, 
    major tick length=2pt, 
    minor tick num=1, 
  ]
    \addplot[color=red, thick] table[x=cache_size, y=average_fitness, col sep=comma]{cache_results.dat};
    \addplot[dashed, black] coordinates {(60,2150) (60,2350)}; 
  \end{axis}
\end{tikzpicture}
\caption{}
\label{fig:cache_plots_avg_fitness}
\end{subfigure}
%
\begin{subfigure}[t]{0.48\columnwidth}
\centering
\begin{tikzpicture}
  \begin{axis}[
    width=\linewidth, 
    height=4.0cm, 
    xlabel={Cache Size ($N$)},
    ylabel={Std. Fitness},
    label style={font=\scriptsize}, 
    tick label style={font=\scriptsize}, 
    grid=major,
    ymin=100, ymax=350, 
    major tick length=2pt, 
    minor tick num=1, 
  ]
    \addplot[color=blue, thick] table[x=cache_size, y=std_fitness, col sep=comma]{cache_results.dat};
    \addplot[dashed, black] coordinates {(60,100) (60,350)}; 
  \end{axis}
\end{tikzpicture}
\caption{}
\label{fig:cache_plots_std_fitness}
\end{subfigure}
\caption{Structural cache-size sensitivity: (a) average fitness score and (b) standard deviation of fitness, both versus cache size. The dashed vertical guide marks the selected cache size, $N=60$.}
\label{fig:combined_cache_plots}
\end{figure}

As shown in Fig.~\ref{fig:cache_plots_avg_fitness}, we observe a rapid improvement in average architectural fitness, which plateaus in the range of $N=12$--$16$. Concurrently, the standard deviation of fitness scores stabilizes in the range of $N=16$--$20$. Consequently, we select $N=60$ for all main experiments. This value provides a safety margin that ensures a dense representation of the Pareto training curriculum, while remaining computationally efficient.

\subsubsection{Component-wise Ablation of Fitness Function}
\label{subsubsec:fitness_ablation}

To validate the individual contributions of the DeepMAD-inspired fitness components defined in Eq.~\ref{eq:fitness_func}, we conducted an ablation study at a constrained computational budget of $\sim$600M MACs on CIFAR-10. We performed independent evolutionary searches ($N=5$) for each configuration and evaluated the mean test accuracy of the discovered subnets. The results are summarized in Table~\ref{tab:ablation_results}.

\paragraph{Impact of Effectiveness ($\rho$)}
The DeepMAD baseline yields an accuracy of 92.60\%. Introducing the effectiveness reward ($\rho$) improves performance by 0.78\%, suggesting that favoring balanced depth-to-width ratios helps identify architectures that perform well in practice.

\paragraph{Impact of Structural Alignment}
Incorporating structural regularizers yields further gains. The Depth Variance Penalty ($Q$), which encourages balanced stage depths, improves accuracy to 93.41\%. Finally, the Non-Decreasing Channel Penalty ($V$), which enforces standard channel expansion conventions, boosts the final accuracy to \textbf{93.62\%} (+1.02\% over the DeepMAD baseline).

Crucially, Table~\ref{tab:ablation_results} highlights that these structural priors act as powerful regularizers. The inclusion of $\rho$, $Q$ and $V$ reduces the performance variance from $\pm$1.14\% to $\pm$0.13\%. The lower variation indicates that the full structural fitness yields more consistent high-performing selections across different search initializations.

\begin{table}[t]
    \centering
    \caption{Ablation of Fitness Components ($\sim$600M MACs, CIFAR-10). Accuracy values are mean $\pm$ standard deviation across five independent evolutionary searches with different initialization seeds.}
    \label{tab:ablation_results}
    \begin{tabularx}{\linewidth}{X c c c}
        \toprule
        \textbf{Configuration} & \textbf{MACs} & \textbf{Acc. (\%)} & \textbf{Gain (pp)} \\
        \midrule
        1. DeepMAD baseline & 599.01 M & 92.60 \scriptsize{$\pm$ 1.14} & - \\
        2. + Effectiveness reward ($\rho$) & 599.53 M & 93.38 \scriptsize{$\pm$ 0.30} & +0.78 \\
        3. + Depth Pen. ($Q$) & 599.62 M & 93.41 \scriptsize{$\pm$ 0.46} & +0.81 \\
        4. + Chan. Pen. ($V$) & 597.80 M & \textbf{93.62} \scriptsize{$\pm$ 0.13} & \textbf{+1.02} \\
        \bottomrule
    \end{tabularx}
\end{table}

\subsubsection{Validation of Predictor-Free Search Hypothesis}
\label{subsubsec:correlation_analysis}

A central hypothesis of DeepFedNAS is that constrained supernet training makes the structural fitness $\mathcal{F}(\mathcal{A})$ positively correlated with validation accuracy. To examine this hypothesis, we evaluate 60 architectures spanning the computational range and measure their accuracy on the fixed validation split.

\begin{figure}[t]
    \centering
    \begin{tikzpicture}
        \begin{axis}[
            width=0.78\linewidth, 
            height=5.2cm,       
            xlabel={Fitness Score},
            ylabel={Validation Accuracy (\%)},
            grid=major,
            grid style={dashed,gray!30},
            label style={font=\small},
            tick label style={font=\footnotesize},
            legend style={font=\footnotesize},
            title style={font=\small, align=center},
            colormap={viridis}{
                rgb255=(68,1,84)
                rgb255=(72,35,116)
                rgb255=(64,67,135)
                rgb255=(52,94,141)
                rgb255=(41,120,142)
                rgb255=(32,144,140)
                rgb255=(34,167,132)
                rgb255=(68,190,112)
                rgb255=(121,209,81)
                rgb255=(189,222,38)
                rgb255=(253,231,36)
            },
            colorbar,
            colorbar style={
                ylabel={MACs (M)},
                yticklabel style={
                    /pgf/number format/fixed,
                    /pgf/number format/precision=0,
                    font=\scriptsize
                },
                ylabel style={font=\scriptsize},
                width=0.2cm, 
            },
            scatter,
            only marks,
            point meta=explicit,
            scatter/use mapped color={
                draw=black!20, 
                fill=mapped color
            },
            mark=*,
            mark size=2.0pt, 
            mark options={fill opacity=0.8},
            enlargelimits=0.05,
            title={\textbf{Fitness vs. Validation Accuracy} \\ Spearman: 0.95, Kendall: 0.83},
            ymin=87, ymax=97,
        ]

        \addplot [
            scatter,
            only marks,
        ] table [
            x=fitness, 
            y=accuracy, 
            meta expr=\thisrow{real_macs}/1000000, 
            col sep=comma
        ] {pareto_fitness_correlation.csv};

        \end{axis}
    \end{tikzpicture}
    \caption{Fitness versus validation accuracy for 60 CIFAR-10 architectures spanning 0.459--3.375~B MACs (Spearman $\rho=0.95$, Kendall $\tau=0.83$). Point color denotes computational cost in millions of MACs.}
    \label{fig:fitness_correlation}
\end{figure}

The strong correlation in Fig.~\ref{fig:fitness_correlation} supports using $\mathcal{F}(\mathcal{A})$ as a structural accuracy proxy for Predictor-Free Search, without fitting an accuracy predictor.

%% file: sections/05_limitations_and_future_work.tex
\section{Limitations and Future Work}
\label{sec:limitations}

Our evaluation is limited to ResNet-style convolutional search spaces and image-classification benchmarks (CIFAR-10, CIFAR-100, CINIC-10). Extending \emph{Pareto-Guided Supernet Training} to Transformers, IoT-specific tasks, and physical hardware platforms such as NVIDIA Jetson, Raspberry Pi, and ARM Cortex-M will require architecture-family-specific search spaces and fitness definitions followed by task-level validation. Quantization, mixed precision, and joint architecture--precision search are also left for future work; such an extension would add precision choices to the architecture genome and require a device-specific resource model.

Consistent with SuperFedNAS \cite{khare2023superfednas}, we assume clients can train their assigned subnets. DeepFedNAS therefore optimizes \emph{deployment} architectures but does not enforce hard client hardware constraints during \emph{training}. A natural extension is system-aware sampling that restricts client-subnet assignments using device capabilities such as available RAM and processor type.

Finally, \emph{Predictor-Free Search} depends on the correlation between structural fitness $\mathcal{F}(\mathcal{A})$ and accuracy. For the 60 CIFAR-10 architectures in Fig.~\ref{fig:fitness_correlation}, the Spearman correlation is $\rho=0.95$. Under substantial domain shift or radically different modalities, learned predictors or task-specific calibration of the structural regularizers may provide more reliable ranking.

%% file: sections/06_conclusion.tex
\section{Conclusion}
\label{sec:conclusion}
DeepFedNAS addresses two bottlenecks in federated architecture adaptation for heterogeneous IoT fleets: unguided supernet training and costly post-training accuracy prediction. Its unified fitness function enables \emph{Federated Pareto Optimal Supernet Training}, which trains from an elite architecture curriculum, and \emph{Predictor-Free Search}, which specializes subnets for target resource budgets in $\sim$20 seconds. In our CIFAR-10 benchmark, this removes the baseline workload of 10,000 subnet evaluations over 5,000 validation images, or 50 million image-level forward evaluations. Across three datasets and diverse non-IID settings, DeepFedNAS achieves state-of-the-art accuracy. On CIFAR-100, its lowest-budget result exceeds SuperFedNAS's best mean accuracy while using $2.95\times$ fewer parameters. These capabilities support rapid, resource-aware deployment without supernet retraining or learned accuracy-predictor construction.


%% file: sections/backmatter.tex
\bibliographystyle{IEEEtran}
\bibliography{references}

\begin{IEEEbiography}[{\includegraphics[width=1in,height=1.25in,clip,keepaspectratio]{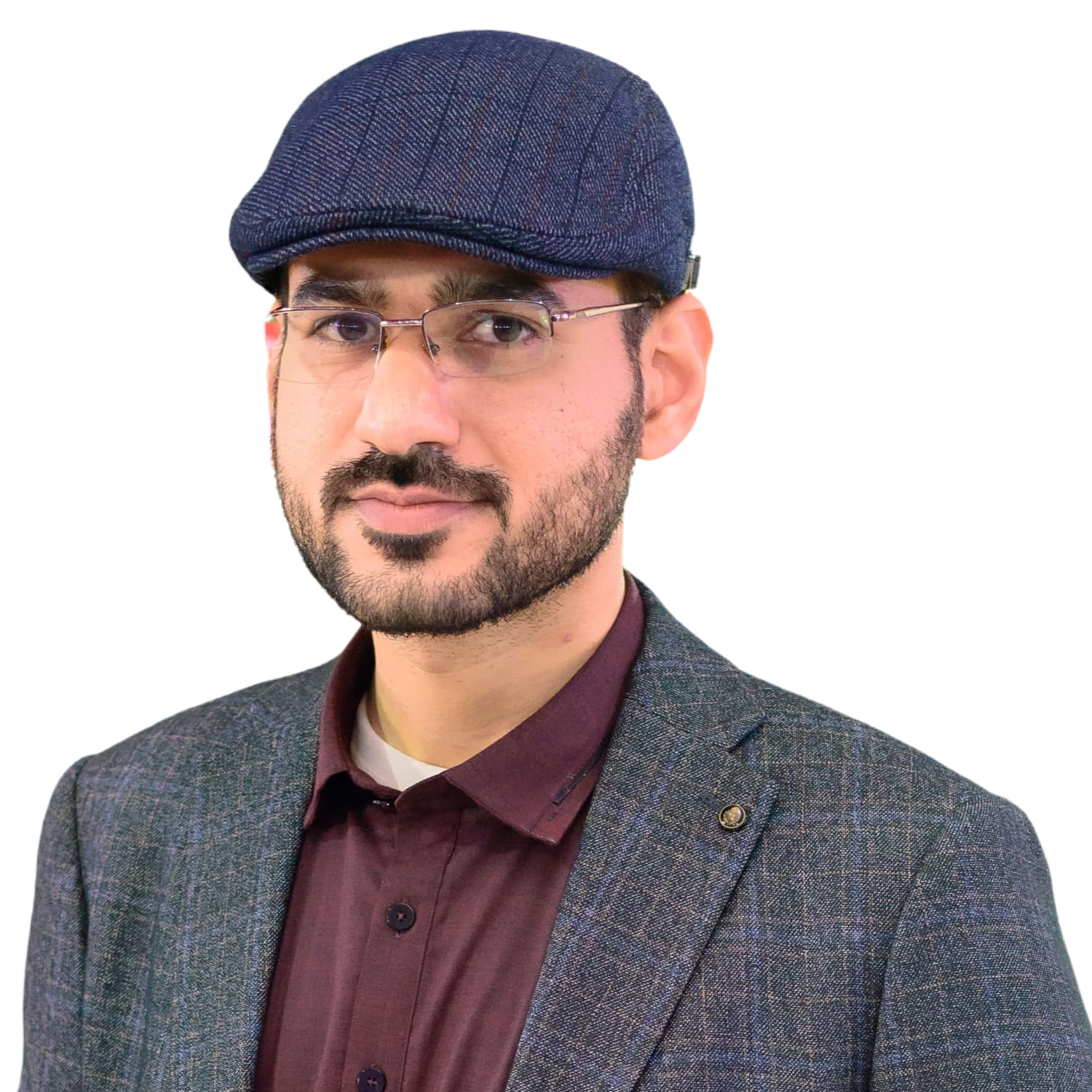}}]{Bostan Khan}
received the B.E.\ degree in Electrical Engineering and the M.S.\ degree in Computer Science from the National University of Sciences and Technology (NUST), Pakistan. He is pursuing the Ph.D.\ degree at M\"{a}lardalen University, Sweden, as part of the AutoFL project. His research focuses on federated learning and neural architecture search for resource-heterogeneous decentralized systems. Previously, he led the AI team at the Machine Vision and Intelligent Systems (MachVIS) Lab at NUST.
\end{IEEEbiography}

\begin{IEEEbiography}[{\includegraphics[width=1in,height=1.25in,clip,keepaspectratio]{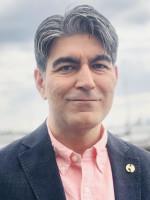}}]{Masoud Daneshtalab}
is a Professor at Mälardalen University (MDU), Sweden, and leads the Heterogeneous System research group (HERO). His research covers centralized and distributed AI, computer vision, biomedical applications, algorithm-hardware co-design, AI acceleration, and federated learning, with emphasis on performance, reliability, security, robustness, and fairness. He has led numerous research projects, authored more than 46 journal and 200 conference papers, and serves on editorial boards and conference committees.
\end{IEEEbiography}

\end{document}